\documentclass[10pt,twocolumn,letterpaper]{article}
\usepackage{amsmath}
\usepackage{wacv} 

\newcommand{\customcellcolor}[1]{%
 \pgfmathsetmacro\colorpercentage{#1 < 80 ? max(min(100.0*(80 - #1)/(80),100.0),0.00) :max(min(100.0*(#1 - 80)/(100-80),100.0),0.00)}%
 \edef\cellcolorcommand{\noexpand\cellcolor{red!\colorpercentage!green!40}}%
 \cellcolorcommand
}

\usepackage{times}
\usepackage{epsfig}
\usepackage{graphicx}
\usepackage{amsmath}
\usepackage{amssymb}
\usepackage{booktabs}
\usepackage[table,xcdraw]{xcolor}
\usepackage{enumitem} 
\usepackage{stmaryrd} 
\usepackage{multirow}
\usepackage{todonotes}
\usepackage{float}
\usepackage{kantlipsum}

\usepackage[pagebackref,breaklinks,colorlinks]{hyperref}
\usepackage{pifont}

\usepackage[capitalize]{cleveref}
\crefname{section}{Sec.}{Secs.}
\Crefname{section}{Section}{Sections}
\Crefname{table}{Table}{Tables}
\crefname{table}{Tab.}{Tabs.}

\newcommand{\Acil}{Incr}
\newcommand{\incracc}{\overline{Acc}}
\newcommand{\arch}{Arch}
\newcommand{\method}{Method}
\newcommand{\tuning}{FT}
\newcommand{\external}{Ext}
\newcommand{\supervision}{Sup}

\newcommand{\methodsp}{\method~}
\newcommand{\tuningsp}{\tuning~}

\newcommand{\supervisionsp}{\supervision~}

\def\wacvPaperID{353} 
\def\confName{WACV}
\def\confYear{2024}

\begin{document}

\title{An Analysis of Initial Training Strategies for Exemplar-Free Class-Incremental Learning}

\author{
Grégoire Petit$^*$\textsuperscript{1,2}, 
Michael Soumm$^*$\textsuperscript{1}, 
Eva Feillet$^*$\textsuperscript{1,4}, 
Adrian Popescu\textsuperscript{1},\\
Bertrand Delezoide\textsuperscript{3},
David Picard\textsuperscript{2},
Céline Hudelot\textsuperscript{4}\\
 \textsuperscript{1}Université Paris-Saclay, CEA, LIST, F-91120, Palaiseau, France\\
 \textsuperscript{2}LIGM, Ecole des Ponts, Univ Gustave Eiffel, CNRS, Marne-la-Vallée, France\\
 \textsuperscript{3}Amanda, 34 Avenue Des Champs Elysées, F-75008, Paris, France\\
 \textsuperscript{4}Université Paris-Saclay, CentraleSupélec, MICS, France\\
{\tt\small g.petit360@gmail.com,} {\tt\small \{michael.soumm, eva.feillet, adrian.popescu\}@cea.fr,}\\
{\tt\small david.picard@enpc.fr,bertrand.delezoide@amanda.com,celine.hudelot@centralesupelec.fr}
}
\maketitle
\thispagestyle{empty}
\begin{abstract}
\vspace{-2pt}
Class-Incremental Learning (CIL) aims to build classification models from data streams. At each step of the CIL process, new classes must be integrated into the model. Due to catastrophic forgetting, CIL is particularly challenging when examples from past classes cannot be stored, the case on which we focus here. To date, most approaches are based exclusively on the target dataset of the CIL process. However, the use of models pre-trained in a self-supervised way on large amounts of data has recently gained momentum. The initial model of the CIL process may only use the first batch of the target dataset, or also use pre-trained weights obtained on an auxiliary dataset. The choice between these two initial learning strategies can significantly influence the performance of the incremental learning model, but has not yet been studied in depth. Performance is also influenced by the choice of the CIL algorithm, the neural architecture, the nature of the target task, the distribution of classes in the stream and the number of examples available for learning. We conduct a comprehensive experimental study to assess the roles of these factors. We present a statistical analysis framework that quantifies the relative contribution of each factor to incremental performance. Our main finding is that the initial training strategy is the dominant factor influencing the average incremental accuracy, but that the choice of CIL algorithm is more important in preventing forgetting. Based on this analysis, we propose practical recommendations for choosing the right initial training strategy for a given incremental learning use case. These recommendations are intended to facilitate the practical deployment of incremental learning.
\end{abstract}
\begin{table}[htbp]
 \centering
 \resizebox{\linewidth}{!}{%
 \begin{tabular}{ccccc||cccccc}
 \multicolumn{5}{c}{\textbf{Initial training strategy}} & \multicolumn{6}{c}{\textbf{CIL Algorithms}} \\
 & & & & & \multicolumn{2}{c}{BSIL~\cite{jodelet2021balanced}} & \multicolumn{2}{c}{DSLDA~\cite{hayes2020_deepslda}} & \multicolumn{2}{c}{FeTrIL~\cite{fetril_Petit_2023_WACV}} \\
 \arch & \method & \tuning & \external & \supervision & $\mu_{\incracc}$ & W & $\mu_{\incracc}$ & W & $\mu_{\incracc}$ & W \\
RN50 & CE & $\checkmark$ & $\times$ & SL & \customcellcolor{ 44.9 } 44.9 & 0 & \customcellcolor{ 53.7 } 53.7 & 4 & \customcellcolor{ 51.0 } 51.0 & 0 \\
RN50 & CE & $\times$ & $ \checkmark $ & SL & \customcellcolor{ 39.9 } 39.9 & 0 & \customcellcolor{ 61.4 } 61.4 & 0 & \customcellcolor{ 60.6 } 60.6 & 0 \\
RN50 & CE & $\checkmark$ & $ \checkmark $ & SL & \customcellcolor{ 62.9 } 62.9 & 1 & \customcellcolor{ 65.3 } 65.3 & 0 & \customcellcolor{ 68.4 } 68.4 & 1 \\
RN50 & BYOL & $\checkmark$ & $\times$ & SSL & \customcellcolor{ 11.2 } 11.2 & 0 & \customcellcolor{ 42.2 } 42.2 & 0 & \customcellcolor{ 34.4 } 34.4 & 0 \\
RN50 & BYOL & $\times$ & $ \checkmark $ & SSL & \customcellcolor{ 35.3 } 35.3 & 0 & \customcellcolor{ 63.3 } 63.3 & 0 & \customcellcolor{ 62.0 } 62.0 & 0 \\
RN50 & BYOL & $\checkmark$ & $ \checkmark $ & SSL & \customcellcolor{ 60.2 } 60.2 & 0 & \customcellcolor{ 70.0 } 70.0 & 2 & \customcellcolor{ 70.2 } 70.2 & 0 \\
RN50 & MoCoV3 & $\checkmark$ & $\times$ & SSL & \customcellcolor{ 14.9 } 14.9 & 0 & \customcellcolor{ 49.6 } 49.6 & 0 & \customcellcolor{ 41.1 } 41.1 & 0 \\
RN50 & MoCoV3 & $\times$ & $ \checkmark $ & SSL & \customcellcolor{ 36.3 } 36.3 & 0 & \customcellcolor{ 67.9 } 67.9 & 1 & \customcellcolor{ 65.3 } 65.3 & 0 \\
RN50 & MoCoV3 & $\checkmark$ & $ \checkmark $ & SSL & \customcellcolor{ 64.7 } 64.7 & 2 & \customcellcolor{ 71.8 } 71.8 & 2 & \customcellcolor{ 72.0 } 72.0 & 0 \\
ViT-S & DeiT & $\times$ & $ \checkmark $ & SL & \customcellcolor{ 35.0 } 35.0 & 0 & \customcellcolor{ 58.7 } 58.7 & 0 & \customcellcolor{ 56.3 } 56.3 & 0 \\
ViT-S & DeiT & $\checkmark$ & $ \checkmark $ & SL & \customcellcolor{ 11.2 } 11.2 & 0 & \customcellcolor{ 37.4 } 37.4 & 0 & \customcellcolor{ 27.4 } 27.4 & 0 \\
ViT-S & DINOv2 & $\times$ & $ \checkmark $ & SSL & \customcellcolor{ 70.4 } 70.4 & 4 & \customcellcolor{ 75.7 } 75.7 & 9 & \customcellcolor{ 72.4 } 72.4 & 6 \\
ViT-S & DINOv2 & $\checkmark$ & $ \checkmark $ & SSL & \customcellcolor{ 24.0 } 24.0 & 0 & \customcellcolor{ 45.9 } 45.9 & 0 & \customcellcolor{ 39.2 } 39.2 & 0 \\
\end{tabular}
}
\caption{
Performance of three EFCIL algorithms with different training strategies for the initial model, averaged over 16 target datasets and two EFCIL scenarios.
BSIL~\cite{jodelet2021balanced} is a recent EFCIL algorithm which is representative of fine-tuning-based CIL works.
DSLDA~\cite{hayes2020_deepslda} and FetrIL~\cite{fetril_Petit_2023_WACV} adapt linear probing~\cite{kumar2022fine} for EFCIL.
We present the averaged incremental accuracy ($\mu_{\incracc}$) and the number of cases (W) in which a combination of algorithm and initial training strategy performs best for a combination of target dataset and EFCIL scenario (see Sec.\ref{sec:experimental_setting}).
Initial training strategies are defined by: \arch - deep architecture used (ResNet50 (RN50)~\cite{he2016_resnet} or vision transformer (ViT-S)~\cite{dosovitskiy2020image}); \methodsp - initial training method; \tuningsp - fine-tuning on initial classes of the target dataset; \external - use of an external dataset, such as ILSVRC~\cite{olga2015_ilsvrc}; \supervisionsp - type of supervision for the initial model: self-supervised (SSL) or supervised (SL).
}
\label{tab:teaser_table} 
\end{table}

\def\thefootnote{*}\footnotetext{denotes equal contribution}\def\thefootnote{\arabic{footnote}}
\section{Introduction}
\label{sec:intro}
Real-world applications of Machine Learning (ML) often involve training models from data streams characterized by distributional changes and limited access to past data~\cite{hayes2022online,van2019three}.
This scenario presents a challenge for standard ML algorithms, as they assume that all training data is available at once.
Continual learning addresses this challenge by building models designed to incorporate new data while preserving previous knowledge~\cite{ring1997child}.
Class-Incremental Learning (CIL) is a type of continual learning that handles the case where the data stream is composed of batches of classes. 
It is particularly challenging in the exemplar-free case (EFCIL), i.e. when storing examples of previous classes is impossible due to memory or confidentiality constraints~\cite{hayes2020_deepslda,zhu2022self}.
CIL algorithms must find a balance between knowledge retention, i.e. stability, and adaptation to new information, i.e. plasticity~\cite{masana2021_study,wu2021striking,mermillod2013_stability_plasticity}.
Many existing EFCIL methods
~\cite{jodelet2021balanced,li2016_lwf,rebuffi2017_icarl,zhu2021class,zhu2021pass,zhu2022self,madaan2023heterogeneous} update the model at each incremental step using supervised fine-tuning combined with a distillation loss, and thus tend to favor plasticity over stability.
Another line of work~\cite{hayes2020_deepslda,fetril_Petit_2023_WACV} freezes the initial model and only updates the classifier.
This approach has recently gained interest~\cite{janson2022simple,pelosin2022simpler,wang2022learning} due to the availability of models pre-trained on large external datasets, often through self-supervision~\cite{he2020momentum,oquab2023dinov2}.
While pre-trained models provide diverse and generic features, there are limits to their transferability~\cite{abnar2021exploring}, and these limits have not been studied in depth in the context of EFCIL. 

We propose a comprehensive analysis framework to disentangle the factors which influence EFCIL performance.
Focus is put on the strategies to obtain the initial model of the incremental process. We consider the type of neural architecture, the training method, the depth of fine-tuning, the availability of external data, and the supervision mode for obtaining this initial model.
The initial training strategies are compared using three EFCIL algorithms, representative for the state of the art, on 16 target datasets, under 2 challenging CIL scenarios.
The obtained results are summarized in Table~\ref{tab:teaser_table}.
The main findings are that: (1) pre-training with external data improves accuracy, (2) self-supervision in the initial step boosts incremental learning, particularly when the pre-trained model is fine-tuned on the initial classes, and (3) EFCIL algorithms based on transfer learning have better performance than their fine-tuning-based counterparts.
However, the distribution of best performance, presented in Table ~\ref{tab:teaser_table}, shows that no combination of an EFCIL algorithm and an initial training strategy is best in all cases. This echoes the results of previous studies such as \cite{belouadah2021_study, Feillet_2023_WACV}. 
Therefore, it is interesting to understand the contribution of the different factors influencing EFCIL performance.
To this aim, we analyze these strategies in depth in Section~\ref{sec:analysis}, and use this analysis to formulate EFCIL-related recommendations in Section~\ref{sec:recommendations}.
The insights brought by the proposed analysis could benefit both continual learning researchers and practitioners.
The proposed framework can improve the evaluation and analysis of EFCIL methods. Continual learning practitioners can use the results of this study to better design their incremental learning systems.

\section{Background}
\label{sec:background}
\vspace{-3pt}
\subsection{Pre-training methods}
\vspace{-3pt}
Transfer learning involves using a model trained on a source dataset as a starting point for training another model on a target dataset~\cite{ribani2019transfer}. 
In the case of transfer learning, the weights of the target model are generally initialized with the weights of the source model. These weights can remain fixed, except for the classification layer (\textit{linear probing}), or they can be updated using the target data (\textit{fine-tuning}).
Transfer learning has several practical advantages~\cite{tan2018survey}. 
It reduces the computational effort to train a new model on a new dataset. It also enables learning an accurate model in few-shot settings, because models pre-trained on large datasets are able to extract complex features even for new input data. 
Some authors investigate how to pre-train the model in order to make it more transferable ~\cite{NEURIPS2018_generalisation,tamaazousti2017manuscript,kornblith18transfer}. 
Model generalization is favored by the quantity, quality, and diversity of its source training data~\cite{oquab2023dinov2}. 
However, the parametric footprint of pre-trained models, typically in the range of hundreds of millions, is often too high for continual learning applications~\cite{hayes2022online}.
Smaller models can be obtained from larger models through knowledge distillation~\cite{hinton2015_distillation,touvron2021deit}.

Self-Supervised Learning (SSL) has recently gained interest thanks to its ability to produce diverse, reusable features for downstream tasks. 
SSL enables a model to learn from unlabeled data without relying on explicit annotations~\cite{jing2020self}. It leverages the inherent structure or information present within the data itself to create surrogate labeling tasks e.g. predicting missing image patches, image rotations, or colorizations. 
For example, MoCov3~\cite{he2020momentum,chen2021empirical} uses a contrastive loss function to obtain similar representations for two randomly augmented crops of the same input image.
Recently, SSL methods trained on large datasets such as BYOL~\cite{grill2020bootstrap} and DINOv2~\cite{oquab2023dinov2} have provided efficient feature extractors, reusable for other tasks. 
We note that, while the reuse of pre-trained models as frozen feature extractors is easy, their fine-tuning in the presence of domain shift may be challenging~\cite{kumar2022fine}.
This is important in the context of CIL since many existing models are based on fine-tuning. 

We compare various pre-training methods to obtain the initial model of a CIL process (Subsec. \ref{subsec:strategies}). We consider (i) the case where the initial model is trained using only the initial batch of target data and (ii) the case where an external dataset was available for pre-training. 
In the first case, the initial model is either obtained using classic supervised learning or using an SSL algorithm, here MoCov3~\cite{chen2021empirical}. 
In the second case, we start the EFCIL process with a model whose weights have been learned either in a supervised manner or in a self-supervised manner~\cite{grill2020bootstrap,oquab2023dinov2} or through distillation~\cite{touvron2021deit}. This allows us to study the transferability of the resulting initial models, in combination with various CIL methods. 

\subsection{Class-Incremental Learning (CIL)}
Continual learning aims to build models that are able to continuously and adaptively learn about their environment.
In CIL, learning a classification model is a sequential process, where each step in the sequence consists of integrating a set of new classes into the model~\cite{belouadah2021_study,lange2019,masana2021_study,parisi2019_continual}. In the exemplar-free setting, at a given stage in the process, the model must be able to recognize all the classes encountered so far, with access only to the current batch of classes. 

The main challenge faced by CIL models is their tendency to forget previously acquired information when confronted with new information. This phenomenon is called \textit{catastrophic forgetting} or \textit{catastrophic interference}, as it is caused by the ``interference" of new information with previous information~\cite{mccloskey:catastrophic,french1999catastrophic}. Forgetting may be reduced by storing examples from past classes, a strategy called \textit{rehearsal}~\cite{rebuffi2017_icarl}.
However, the availability of past data and the possibility to store it may be unrealistic in practice. 
Thus, we focus on EFCIL rather than rehearsal-based CIL. 

Two main directions may be considered to deal with forgetting in artificial neural networks. 
A first family of CIL approaches lets the network grow as new capabilities must be learned, e.g.~\cite{wang2017_growingabrain}. 
In the extreme case, this approach can result in zero forgetting. But at the same time, it is not realistic to make the model grow infinitely. 
A second family of methods considers a network of constant size throughout the incremental process (except the classifier) and proposes various strategies for obtaining models which ensure a balance between stability, i.e. preserving the performance for past classes, and plasticity, i.e. learning to recognize new classes.
The weights of the initial model may be fine-tuned in combination with knowledge distillation between the previous model and the model which is currently learned~\cite{li2016_lwf,hou2019_lucir,jodelet2021balanced,douillard2020podnet,zhu2021class,zhu2021pass,zhu2022self}.
This type of approach favors plasticity over stability because at each incremental step, all model weights are updated using the training images of the latest classes. 
An alternative is to use a pre-trained model or to freeze the model learned in the initial incremental state and to train only a linear classification layer afterward~\cite{belouadah2018_deesil,hayes2020_deepslda,fetril_Petit_2023_WACV}.
Recent works propose to use a large pre-trained model combined with a k-NN classifier as a challenging baseline for continual learning algorithms~\cite{janson2022simple,pelosin2022simpler}. 
These works adapt linear probing to an incremental context. 
They favor stability over plasticity since the feature extractor is not adapted during the incremental process~\cite{masana2021_study}.
The main challenge lies in the transferability of the initial feature extractor to new classes. 
Importantly, they are much faster to train than the fine-tuning-based methods because only the classification layer is updated.

Recent works propose to improve the learned representation by fine-tuning the model with a combination between cross-entropy loss and a self-supervised learning objective~\cite{fini2022self,tang2023kaizen}. 
Another recent trend in CIL is to use a pre-trained model as an efficient starting point for the incremental process~\cite{TIAN2023transfer,wu2022strong_pretrained}. 
The authors of~\cite{gallardo2021selfsup} explore the use of a fixed feature extractor pre-trained in a self-supervised way. 
The authors of~\cite{wang2022learning} also propose a method based on dynamic prompting. 
In~\cite{ostapenko22afoundation}, pre-trained models are used to propose a compute-low method with a replay of past training samples. 
From a practical perspective, using a pre-trained feature extractor is also interesting for cases where training data is scarce, as in few-shot CIL~\cite{ahmad2022few}.

The present work proposes a comprehensive study of training strategies for the initial model, with a focus on the interaction of these methods with different EFCIL algorithms. 
We experiment with transformer-based and CNN architectures, in combination with fine-tuning-based and transfer-learning-based EFCIL algorithms.

\section{Problem statement}
\label{sec:pb_statement}

\begin{figure}[hbt]
 \centering
 \includegraphics[width=0.37\textwidth]{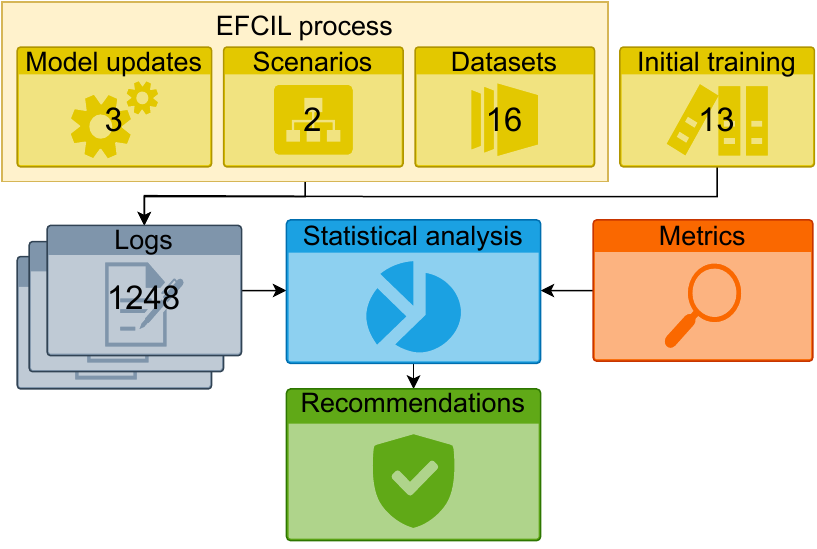}
 \caption{Overview of the proposed analysis framework of initial training strategies for EFCIL.}
 \label{fig:pretril_methodology}
 \vspace{-10pt}
\end{figure}
\vspace{-3pt}
We summarize our proposed analysis framework in Figure~\ref{fig:pretril_methodology}.
It combines a comprehensive modeling of the EFCIL process and initial training strategies as inputs for a statistical analysis that uses different EFCIL metrics.
Recommendations for the design of EFCIL approaches are made based on the conclusions of the statistical analysis.

\subsection{EFCIL process}
\label{subsec:efcil_definition}
Let us consider a dataset $\mathcal{D}$ split over $K$ subsets, $\mathcal{D} = \mathcal{D}_1 \cup \mathcal{D}_2 \cup \dots \cup \mathcal{D}_K$,
and an EFCIL algorithm $\Acil$. A CIL process consists in learning a classification model sequentially over $K$ non-overlapping steps using $\Acil$. At each step $k \in \llbracket 1,K \rrbracket$, the model is updated using $\Acil$ and the data subset $\mathcal{D}_k$, whose associated set of classes is denoted by $\mathcal{C}_k$.
The data subsets $\mathcal{D}_1, \mathcal{D}_2, \cdots, \mathcal{D}_K$ composing the complete dataset $\mathcal{D}$ satisfy the following constraint: for $k,k' \in \{1,2,\dots,K\}$ with $k \neq k'$, $\mathcal{C}_{k} \cap \mathcal{C}_{k'} = \emptyset$, i.e. each class is only present in a single data subset.
The use of an exemplar-free algorithm $\Acil$ implies that when the training is performed at the $k^{th}$ step, no example from any of the data subsets of the previous steps can be accessed. 
Although this is a more difficult setting, it is also more realistic in practice~\cite{hayes2022online, belouadah2021_study}.

\textbf{Incremental model updates.} The initial model $\mathcal{M}_1$ is obtained following one of the training strategies presented in \ref{subsec:strategies}. At the $k^{th}$ step of the CIL process, $k \in \llbracket2, K \rrbracket$, the classification model $\mathcal{M}_k$ recovers the weights of the model $\mathcal{M}_{k-1}$ obtained in step $k-1$ and is updated using the data subset $\mathcal{D}_k$ and the algorithm $\Acil$. 
Many EFCIL algorithms~\cite{jodelet2021balanced} fine-tune all network weights at each incremental step, thus favoring plasticity. Alternatively, algorithms such as~\cite{hayes2020_deepslda,fetril_Petit_2023_WACV} only retrain the classifier, thus favoring stability. 
As a compromise, it is also possible to freeze a part of the model and to update only the last layers.
We cover these three cases in our experiments. 

\textbf{Scenario.} 
A CIL scenario is characterized by the distribution of classes among the steps of the CIL process. 
We denote by $b$ the proportion of the classes available in the initial step: $b=Card(\mathcal{C}_1)/Card(\mathcal{C})$. 
There are two commonly used scenarios~\cite{belouadah2021_study} (i) equal splitting of classes across the steps or (ii) half of the classes in the first step and the rest of the classes are divided equally between subsequent steps.

\subsection{Training strategies for the initial model}
\label{subsec:strategies}
In the following, we describe the main characteristics of the training strategies used in our experimental study to obtain the initial model of the incremental learning process. Further experimental settings are reported in Section~\ref{sec:experimental_setting}. 

\textbf{Network architecture.}
So far, most CIL methods have been proposed in combination with a convolutional neural network, but visual transformer (ViT) networks have recently gained popularity in CIL~\cite{Douillard_2022_Dytox}. In order to provide a fair comparison between the two types of architecture, we use a ResNet50~\cite{he2016_resnet}, and a ViT-Small~\cite{dosovitskiy2020image} network, which have a close number of parameters (23.5M and 22.1M parameters respectively).

\textbf{Model initialization.}
At the first step of the CIL process, the weights of the model may either be randomly initialized or transferred from a pre-trained model.
In the second case, depending on the choice of the user, the dataset $\mathcal{D}^\ast$ used for pre-training may either be an auxiliary dataset (e.g. ILSVRC~\cite{olga2015_ilsvrc}), referred to as \textit{source} dataset, or the first data subset $\mathcal{D}_1$ of the incremental process. 

\textbf{Label availability.}
We consider that all examples from the target dataset $\mathcal{D}$ are labeled, and we experiment with both supervised learning and self-supervised learning to obtain the initial model using $D_1$. 
Labels may not be available for the external dataset $\mathcal{D}^\ast$. In this case, the training initialization is performed using a self-supervised pre-training algorithm (e.g., DINOv2~\cite{oquab2023dinov2}).

\section{Experimental setting}
\label{sec:experimental_setting}
We describe the experimental parameters and the metrics we use to evaluate EFCIL models. 
The combination of parameters results in 1,248 experiments in total (Figure \ref{fig:pretril_methodology}).

\subsection{Initial training strategies}
\vspace{-3pt}
We compare different strategies for training an initial model, as summarized in Table~\ref{tab:teaser_table}.
We use Resnet50~\cite{he2016_resnet} and ViT-S~\cite{dosovitskiy2020image} networks, which are representative of CNNs and transformers and have similar sizes. 
The training is done either using a self-supervised method (BYOL~\cite{grill2020bootstrap}, DINOv2~\cite{he2020momentum}, MoCov3~\cite{chen2021empirical}) or a supervised one (DeiT, cross-entropy (CE)). 
We present results for pre-training with external data (i.e. ILSVRC~\cite{olga2015_ilsvrc} for BYOL, DeiT and CE; a 150M-images dataset + ILSVRC for DINOv2) and training on the first batch.
We compare the effect of (i) freezing the weights of the pre-trained model or (ii) further optimizing the last layers of the model (e.g. the last convolutional block in ResNet50) on the initial data subset $\mathcal{D}_1$. 
The first type of experiment is denoted by the suffix ``\textit{-t}" (transfer), the second by the suffix ``\textit{-ft}" (fine-tuning). In the case where the pre-training algorithm is applied to $\mathcal{D}_1$ and not to $\mathcal{D}^\star$, there is no suffix.
\subsection{Target datasets} 
\vspace{-3pt}
For a comprehensive evaluation and to account for the diversity of visual tasks, we evaluate the training strategies on 16 target datasets, sampled from publicly available datasets. 
They cover different domains (plants, animals, landmarks, food, faces, traffic signs etc.), and different types of images (natural, drawings, paintings). 
IMN100$_{1}$ and IMN100$_{2}$ consist of 100 classes randomly selected from ImageNet-21k~\cite{deng2009_imagenet}. 
Flora is a thematic subset of ImageNet consisting of 100 classes belonging to the ``flora" concept.
IMN100$_{1}$, IMN100$_{2}$ and Flora have no mutual overlap and no overlap with ILSVRC~\cite{deng2009_imagenet,olga2015_ilsvrc}. 
Amph100 and Fungi100, sampled from iNaturalist~\cite{van2018inaturalist}, respectively contain 100 classes of amphibians and fungi, selected so as to avoid overlap with animal and fungi classes from ILSVRC. 
We also sample 100-class subsets from other popular datasets: WikiArt100~\cite{hugganwikiart2022}, Casia100~\cite{yi2014casia}, Food100~\cite{bossard2014_food101}, Air100~\cite{maji2013aircraft}, MTSD100~\cite{madani2016malaysian}, Land100~\cite{weyand2020landmarksv2}, Logo100~\cite{Wang2020Logo2K} and Qdraw100~\cite{ha2017quickdraw}. 
Finally, we consider three 1000-class subsets: Casia1k~\cite{yi2014casia}, Land1k~\cite{noh2017landmarksv1}, and iNat1k~\cite{van2018inaturalist}. 
The number of training images per dataset varies from 60 to 750. 
More details on the datasets are provided in the supplementary material. 

\subsection{Incremental learning}
\label{subsec:IL_settings}
\vspace{-3pt}
\textbf{EFCIL scenario $b$.} 
We experiment on two widely used CIL scenarios~\cite{hou2019_lucir,belouadah2021_study}. In the first scenario, the classes are equally distributed over 10 steps, e.g. 10 classes per step for a 100-class dataset. In the second scenario, half of the classes are learned in the initial step, and the other half is equally distributed over 10 incremental steps, e.g. $50 + 10 \cdot 5$ classes for a 100-class dataset. 

\textbf{CIL algorithm $\Acil$.} We experiment with one fine-tuning based algorithm, namely BSIL\cite{jodelet2021balanced}, which adds a balanced softmax without exemplars to LUCIR~\cite{hou2019_lucir}. We also experiment with two fixed-representation-based algorithms, namely DSLDA~\cite{hayes2020_deepslda} and FeTrIL~\cite{fetril_Petit_2023_WACV}. 

\subsection{Metrics} 
\label{subsec:metrics}
The performance of EFCIL models can be evaluated in several ways~\cite{masana2021_study}, discussed below.

\noindent 
\textbf{Average incremental accuracy $\incracc$.} In EFCIL, a model trained over a $K$-step incremental process is commonly evaluated using the average incremental accuracy~\cite{zhu2021pass,zhu2022self,zhu2021class,jodelet2021balanced}. 
We denote it by $\incracc$ and compute it by:
\begin{equation}
\label{eq:avg_incr_acc}
 \incracc~=~\frac{1}{K-1} \sum_{k=2}^K acc(\mathcal{M}_k, \bigcup_{i=1}^k \mathcal{D}_i)
\end{equation} 
where $acc(\mathcal{M}, D)$ is the accuracy of the model $\mathcal{M}$ on the dataset $D$. 
Following common practice in CIL~\cite{castro2018_e2eil,fetril_Petit_2023_WACV,zhu2022self}, $\incracc$ does not take the accuracy of the initial model into account.

\noindent \textbf{Average forgetting $F$.} Average forgetting, denoted here by $F$, is computed by: 
\begin{equation}
\label{eq:avg_forgetting}
 F~= b\times f(\mathcal{D}_1) + \frac{1-b}{K-1} \sum_{k=2}^{K} f(\mathcal{D}_k)
\end{equation}
where $f(\mathcal{D}_{k}) = \max\limits_{k' \in \llbracket k, K\rrbracket} acc(\mathcal{M}_{k'}, \mathcal{D}_k)-acc(\mathcal{M}_K, \mathcal{D}_{k}))$ is the difference between the best performance achieved on the data subset $\mathcal{D}_k$ during the EFCIL process and the final performance of the model on this data subset~\cite{mirzadeh2022wide}. 

\noindent \textbf{Initial accuracy ${Acc}_1$.} To unskew the statistical models we present in Section \ref{sec:analysis}, we consider the initial accuracy, defined as the accuracy of the first model on the first data subset $\mathcal{D}_1$ and denoted by ${Acc}_1$, i.e. ${Acc}_1 = acc(\mathcal{M}_1, \mathcal{D}_1)$. 

\noindent \textbf{Final accuracy ${Acc}_K$.} The accuracy of the last model of the incremental learning process on the complete dataset $\mathcal{D}$ is denoted by ${Acc}_K$, i.e. ${Acc}_K = acc(\mathcal{M}_K, \mathcal{D})$.

$\incracc$ gives more weight to past classes since at each step, the model is evaluated on all classes seen so far. Consequently, a high average incremental accuracy does not guarantee a high accuracy on the latest classes, particularly when half of the classes are learned initially.
Forgetting is complementary to accuracy, as it focuses on model stability. A low value for $F$ indicates that, on average, the performance for a given class remains stable over the incremental process. 

\section{Analysis of results}
\label{sec:analysis}

We present a statistical analysis of the results from Table~\ref{tab:teaser_table}, which highlights the effects of pre-training strategies and of EFCIL algorithms on EFCIL performance.
The statistical model and associated findings are presented below.

\subsection{Modeling causal effects}
\label{subsec:causal}
Our objective is to identify the primary factors that influence the performance of EFCIL algorithms. 
To interpret causal effects, we employ multiple linear regressions using the Ordinary Least Squares (OLS) method, following common statistical and econometric practices~\cite{angrist2009mostly,gareth2013introduction}.
In a linear regression, we aim to explain a target variable $Y$ using explanatory variables $X_i$. The target variable is said endogenous, i.e. determined by its relationship with other variables. If the outcome of a variable $X_i$ is selected by the experimenter, it is said to be exogenous, i.e. not caused by other variables.
For a given experiment, we denote by $Y$ the target metric accuracy (endogenous), $Data$ the evaluation dataset (exogenous), $Train$ the initial training strategy (exogenous), and $Incr$ the incremental algorithm (exogenous). 
We also consider the initial accuracy $Acc_1$ as an endogenous variable that may influence performance and can be controlled in our regressions. 
Other parameters, such as the total number of classes or the dataset, are examined as potential predictors of a metric.

An OLS regression fits a model of the following form:
\begin{equation}
\label{eq:OLS}
Y = \beta_0 + \beta_1 Train +\beta_2 Incr +\beta_3 Data + \ldots +\varepsilon,
\end{equation}
where the intercept $\beta_0$ is a scalar and $\varepsilon$ is assumed to be normally distributed Gaussian noise.
Since $Train$, $Incr$, and $Data$ are categorical, we encode them as one-hot vectors. Thus, $\beta_1$, $\beta_2$, and $\beta_3$ are vectors of the same size as the number of possible categories for each variable. To emphasize the explanatory variables and to simplify notation, in the following we denote the above regression model (Eq.~\ref{eq:OLS}) as ``$Y \sim Train + Incr + Data +\ldots$".

Under appropriate assumptions\footnote{Primarily, non-perfect collinearity among exogenous variables and the normality of the estimated residuals $\hat{\varepsilon}$}, the estimated coefficients can be interpreted as estimated causal effects. 
The statistical significance of these effects is assessed by examining the \textit{p-value} of the associated Student $t$-test for each coefficient~\cite{gareth2013introduction}.
Following established statistical practices~\cite{gareth2013introduction}, we set the significance value at .05. 
The significance, sign, magnitude, and interpretation of each estimated coefficient depend on the regression model. 
In particular, introducing more exogenous variables can cause instability in the regression. Therefore, for each metric $Y$, we adopt the following methodology to select only the most influential factors:
\vspace{-10pt}
\begin{enumerate}[noitemsep,wide]
\item We use multiple regression models to represent the evaluation metric $Y$ as a linear combination of different variables, or of the product of these variables. We ensure that the chosen regressions exhibit no collinearity or numerical issues\footnote{We assess this by examining the smallest eigenvalue of the Gram matrix of the data $X^TX$. Although Ridge or Lasso regression could address these concerns, their coefficients are less interpretable than those of OLS.}.
\item Subsequently, we select a regression model using the Akaike Information Criterion (AIC)~\cite{Akaike1998}, which regularizes the likelihood of the model based on its degrees of freedom.
\item We interpret the regression coefficients, the coefficient of determination $R^2$, and examine the Q-Q plot of the residuals $\hat{\varepsilon}$ to verify their normality.
\item Next, we conduct an Analysis of Variance (ANOVA)~\cite{gareth2013introduction} on the regression to obtain aggregated statistics on the categorical variables.
\item Finally, we interpret the partial $\eta^2$ derived from the ANOVA as a measure of the importance of each variable.
\end{enumerate}

A regression on a categorical variable requires the setting of a reference value for it. 
Therefore, the coefficient(s) associated with this categorical variable represent the causal effects of this variable \textit{with respect to the reference level}.
However, we want to compare all initial training strategies with each other to derive practical recommendations. Therefore, we use the following protocol to generate pairwise significant differences: 
 (1) perform the same regression multiple times using a different reference category; (2) sum-up the pairwise comparisons in a double-entry matrix;
 (3) since we are performing multiple tests, we need to adjust the significance threshold of each test using Bonferroni correction~\cite{gareth2013introduction}, which consists of dividing the \textit{p-value} threshold by the number of tests;
 (4) plot a heatmap of the pairwise comparisons between the choice of a parameter.

\subsection{Metrics and confounding Factors}
\begin{figure}[h]
 \centering
 \includegraphics[width=0.65\linewidth]{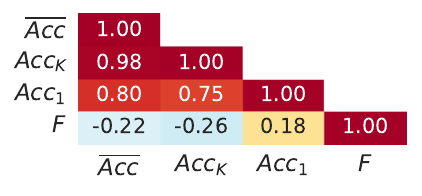}
 \vspace{-10pt}
 \caption{Correlation between the endogenous variables. }
 \vspace{-10pt}
 \label{fig:correlation}
\end{figure}

In Figure~\ref{fig:correlation}, we examine the relationship between the evaluation metrics defined in Subsection~\ref{subsec:metrics}.
We observe a strong positive correlation between $\incracc$ and ${Acc}_K$. There is a weak negative correlation between average incremental accuracy and forgetting, which is expected due to the inherent trade-off between stability (i.e. low forgetting) and plasticity in CIL (i.e. high performance on new classes). 
We note a significant correlation between average incremental accuracy and accuracy in the initial state. 
This correlation is expected since half of our experiments are done with half of the classes in the initial step. 
Additionally, the average incremental accuracy (Eq.~\ref{eq:avg_incr_acc}) evaluates each model on each class, from the first occurrence of the class to the end of the incremental process, thus giving greater influence to earlier classes. 
Conversely, there is a weak correlation between forgetting and initial accuracy. 
This implies that the performance on the initial batch of classes does not significantly impact the model's stability throughout the incremental steps.
\vspace{-3pt}

Based on these observations, we choose the average incremental accuracy $\incracc$ and the average forgetting $F$ as the metrics of interest for our study, and include the effect of the initial accuracy in their models. 
Controlling the initial accuracy in a regression model is important to draw accurate conclusions: if pure accuracy is sought, then it can be left out of the model. However, the goal of CIL algorithms is not solely to be accurate on average, but rather to be accurate while preventing forgetting. Hence, to analyse the actual incremental contribution of each method, initial accuracy should be included in the regression.

\subsection{Factors influencing incremental performance}
This subsection presents the aggregated influence of the considered parameters.
The models and findings presented in Table~\ref{tab:ANOVA} are obtained with the methodology presented in Subsection~\ref{subsec:causal}.
More details on the obtained models can be found in the supplementary material.
\vspace{-5pt}
\begin{table}[htbp]
 \centering
 \resizebox{0.9\linewidth}{!}{%
\small 
\begin{tabular}{c||c|c|c}

Model & $R^2$ & variable & $\eta^2$ \\ \hline \hline
\multirow{3}{*}{$\incracc\sim Incr+Train+Data$} & \multirow{3}{*}{0.69} & \footnotesize$Train$ & 0.32 \\ 
 & & $Data$ & 0.24\\ 
 & & $Incr$ & 0.11 \\ \hline 
\multirow{4}{*}{$\incracc\sim Acc_1+Incr+Train+Data$} & \multirow{4}{*}{0.81} & $Acc_1$ & 0.25\\ 
 & & \footnotesize$Incr$ & 0.22 \\ 
 & & \footnotesize$Train$ & 0.10 \\ 
 & & \footnotesize$Data$ & 0.06 \\ \hline 
\multirow{3}{*}{$F\sim Incr+Train+Data$} & \multirow{3}{*}{0.71} & \footnotesize$Incr$ & 0.61 \\ 
 & & \footnotesize$Train$ & 0.06 \\ \
 & & \footnotesize$Data$ & 0.03 \\
\end{tabular}}
\caption{ANOVA results for each considered regression. Variables are significant at $p<0.05$ and ordered by decreasing importance.}
\label{tab:ANOVA}
\end{table}

\textbf{Main influences}: In Table \ref{tab:ANOVA}, the most significant factor affecting average incremental accuracy is the choice of initial training strategy. However, upon controlling the impact of initial accuracy, the selected incremental algorithm has a greater importance. This distinction is primarily attributed to BSIL, which exhibits an incremental accuracy 16 points below that of FeTrIL and DSLDA on average.

Regarding forgetting, the incremental algorithm is the most influential parameter. Here, this effect is not driven by any specific outlier method. Further analysis shows that initial accuracy also plays a significant role in predicting the level of forgetting. The associated regression coefficient is $.16$ ($\pm .02$), indicating that a 1-point increase in initial accuracy results in a 16-point increase of forgetting. 

Given that accuracy ranges between 0 and 1, a lower initial accuracy decreases the likelihood of experiencing high levels of forgetting.
Hence, a trade-off arises concerning the initial accuracy: while its enhancement greatly improves the average incremental accuracy, it also appears to amplify forgetting. This should be taken into account when comparing CIL algorithms. 
From a research perspective, the incremental algorithm remains influential in the metrics, particularly when controlling for initial accuracy or focusing on forgetting. However, in practical applications of CIL, the final accuracy may be more important. Given its strong correlation with average incremental accuracy, increasing the initial accuracy becomes more advantageous in this case.

\subsection{Comparison of initial training strategies} 
\begin{figure}
 \centering
 \includegraphics[width=\linewidth]{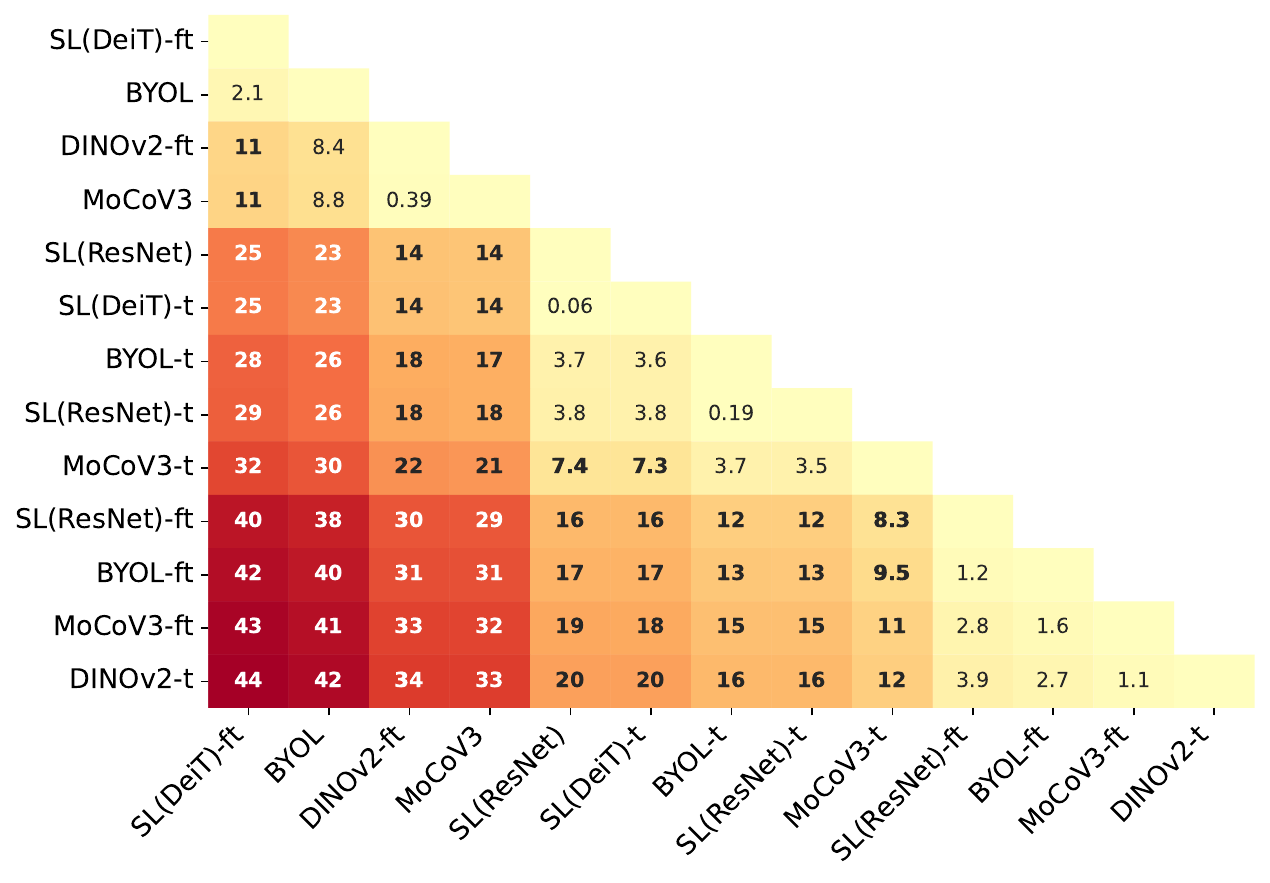}
 \vspace{-10pt}
 \caption{Accuracy gain by using strategy in row $i$ over strategy in column $j$ , e.g. ``The accuracy of BYOL-ft is 17pts higher than SL(ResNet)". Only results in \textbf{bold} are statistically different. 
 }
 \label{fig:pre-train_on_A}
  \vspace{-10pt}
\end{figure}

In Figure \ref{fig:pre-train_on_A}, we observe notable variations in accuracy among different initial training strategies, thus prompting the identification of three regimes:
\vspace{-10pt}
\begin{enumerate}[noitemsep,wide,labelwidth=!]
\item \textbf{Strategies that surpass supervised learning without transfer:} MoCoV3-ft, DINOv2-t, BYOL-ft, SL(ResNet)-ft, MoCov3-t. These approaches exhibit superior performance by generating a robust latent space, whose features are transferable. 
MoCoV3-ft 
enhances its latent space by fine-tuning, enabling better generalization compared to other methods. 
DINOv2-t follows, leveraging its extensive self-supervised training on a very large amount of data. 
BYOL-ft and SL(ResNet)-ft closely follow, highlighting the advantage gained from additional adaptation steps on the target dataset following pre-training. MoCov3-t is fifth, showing that features generated through an adapted self-supervised method have a generalization capability that can be leveraged in CIL.
\item \textbf{Strategies that exhibit no significant improvement over supervised learning without transfer:} SL(ResNet)-ft, BYOL-t, SL(DeiT)-t. Our analysis underlines the capability of well-designed self-supervised methods to outperform supervised pre-training approaches.
\item \textbf{Strategies that underperform compared to supervised learning without transfer:} MoCoV3, BYOL, DINOv2-ft, SL(DeiT)-ft. The inferior performance of self-supervised methods can be attributed to the limited initial data.
Furthermore, the challenging nature of fine-tuning for transformer models contributes to the underwhelming outcomes observed in these models.
\end{enumerate}

The analysis of the average forgetting, illustrated in Figure 3 of the supplementary material, indicates that the majority of pairwise initial training strategies exhibit no significant distinctions.
However,
DINOv2-t exhibits lower forgetting compared to other strategies, including SL (ResNet). This is particularly remarkable considering that DINOv2-t has the highest initial accuracy. Conversely, fine-tuned transfer models (DINOv2-ft, SL(DeiT)-ft) also display a lower forgetting, albeit primarily attributed to their inherently low initial accuracy, which leaves little room for further decline in their accuracy.

\subsection{Further analysis of initial training strategies}
We now inquire whether the preceding general analysis can be nuanced in specific scenarios. To this end, we perform the same analysis as in the previous section by performing the regression on subsets of the data. All complementary graphs that justify the following statements can be found in the supplementary material.

\textbf{Influence of the dataset.} Regarding target datasets that are furthest from the pre-training dataset, the benefit of pre-training with or without fine-tuning is lower due to the domain gap. We note that specialized datasets, such as Qdraw100 and Casia100, also contain smaller images than those of ILSVRC. Whether the difference in performance is caused by a semantic gap or an image-size gap is unclear.

\textbf{Influence of the incremental scenario.} Regarding accuracy, we find that most differences among methods come from the scenarios with 50 initial classes or less. With 10 initial classes, all strategies that were previously not significantly better than SL(ResNet) start to outperform it. In scenarios with 50 initial classes, it becomes more difficult to precisely rank the top initial training strategies. In scenarios with 100 initial classes, no strategy is significantly better than any other one (which can come from the lower number of experiments with these scenarios).

\textbf{Influence of Incremental method.} We find that FeTrIL and DSLDA exhibit a similar pattern for $\incracc$ and $F$, contrary to BSIL. 
For FeTrIL and DSLDA, the differences between the best initial training strategies are less clear, but the general trend previously described still holds, in particular for the accuracy. 
The choice of the training strategy does not clearly impact the forgetting. 
On the other hand, BSIL is much more sensitive to the initial training strategy. Fine-tuned methods clearly outperform classical learning and plain transfer (except for DINOv2-t), whether it concerns the accuracy or the forgetting. Moreover, SL(ResNet) is a stronger baseline for BSIL than for the other methods when considering incremental accuracy.

\begin{figure}[ht]
 \includegraphics[width=\linewidth]{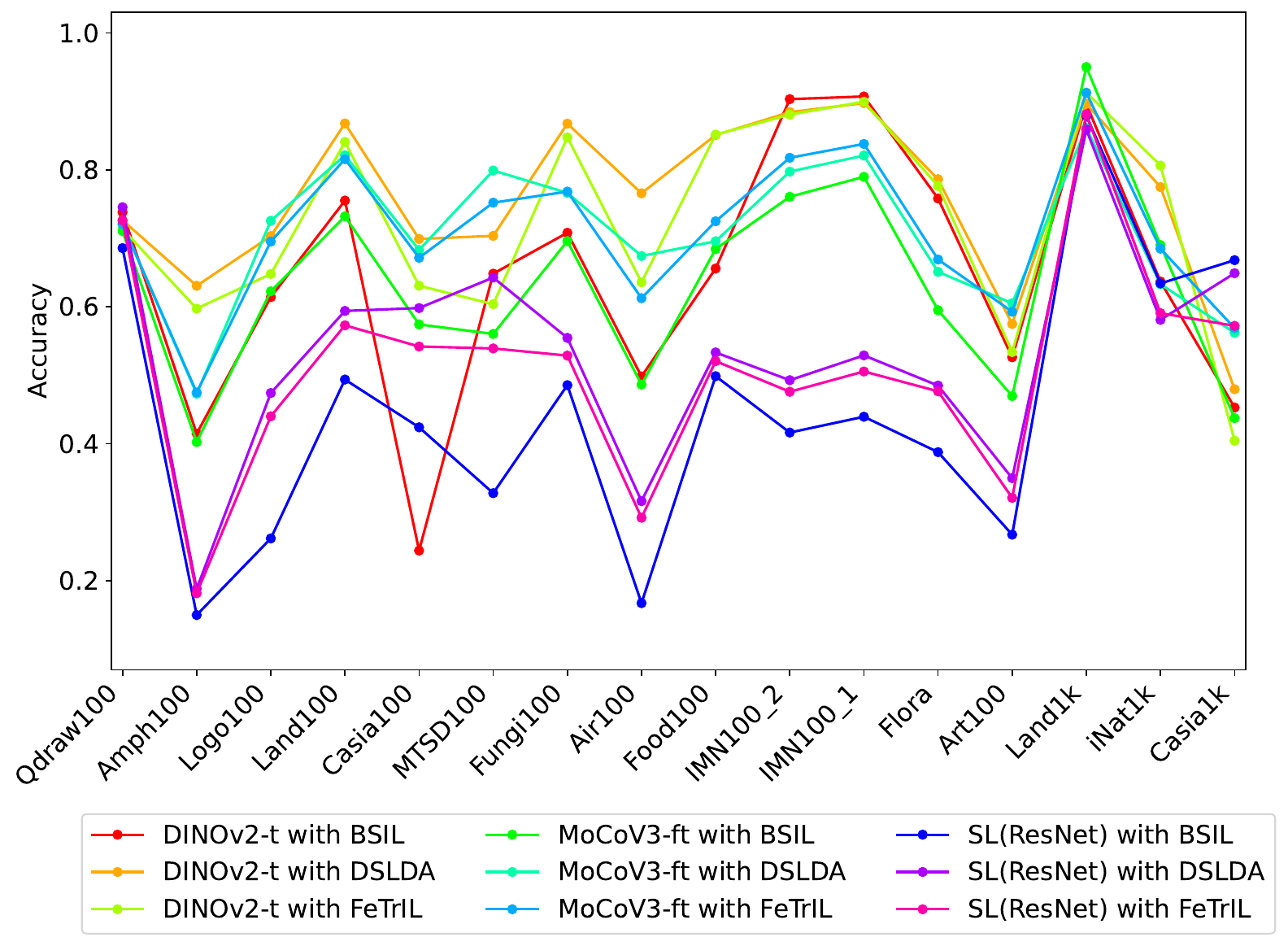}
  \vspace{-10pt}
 \caption{Interaction plot of the best strategies for different transfer types and for the 3 CIL algorithms. Similar slopes indicate similar behaviors. A change in slope indicates a change in behavior.}
 \label{fig:detailed}
\end{figure}
\vspace{-3mm}

\section{Discussion} 
\label{sec:recommendations}
We summarize our findings and propose recommendations for the design of EFCIL approaches.

\textbf{Does the use of a model pre-trained on an external dataset $\mathcal{D}^\star$ always improve performance on the target dataset $\mathcal{D}$?}
Figure~\ref{fig:detailed} highlights that no single initial training strategy outperforms the others on all datasets.
As illustrated in Table~\ref{tab:teaser_table}, pre-training is clearly better on average, but there are exceptions.
Intuitively, the use of a pre-trained model without fine-tuning (DINOv2-t in Figure~\ref{fig:detailed}), is clearly preferable for datasets such as IMN100$_1$ and Flora which are closely related to the dataset used for pre-training. 
Inversely, the supervised training method SL(ResNet) is better when the gap between the source and the target datasets is important, such is the case for Casia1k.
MoCov3-ft is a good compromise since it leverages pre-training, but adapts the representation via partial fine-tuning.
The initial training strategy should be selected by considering characteristics of the dataset such as: number of classes, number of samples per class, domain gap with pre-training, and size of the initial batch of classes. 

\textbf{In the absence of an external dataset, is it better to train the initial model in a supervised way or with a self-supervised learning method?} 
As shown in Figure~\ref{fig:pre-train_on_A},
 supervised learning on the initial data is better on average. 
However, self-supervised learning is better when the amount of data available initially is limited, making it difficult to train a supervised model effectively. 


\textbf{Should the pre-trained model be fine-tuned on the first batch of data, or frozen?}
Existing EFCIL works that use pre-trained transformers keep their weights fixed~\cite{janson2022simple,pelosin2022simpler,wang2022learning}. 
This might be explained by the fact that fine-tuning these models might be detrimental in transfer learning~\cite{kumar2022fine}. 
Inversely, the performance of CNN-based training strategies, such as BYOL or MoCov3, increases after partial fine-tuning.
This is explained by the fact that the layers of CNNs are reusable across tasks, while fine-tuning the last layers with initial target data improves transferability in subsequent EFCIL steps. 


\textbf{How does the performance of EFCIL algorithms vary with initial training strategies?}
Table~\ref{tab:teaser_table} and Figure~\ref{fig:detailed} show that the performance of BSIL varies much more than that of DSLDA and FeTrIL.
This is particularly clear for transformer models, where BSIL performance is strongly degraded when fine-tuning of pre-trained models is used. 
In contrast, the variation of performance for DSLDA and FeTrIL is much lower when testing partial fine-tuning and transfer strategies on top of pre-trained models.
This suggests that both initial training strategies are usable in practice for transfer-learning based EFCIL algorithms. 


\textbf{What is the impact of using transformers versus convolutional neural networks?}
The averaged results presented in Table~\ref{tab:teaser_table} and the detailed ones from Figure~\ref{fig:detailed} show that the difference between the best training strategies based on transformers and on CNNs is small. 
This is particularly the case when CNNs are pre-trained in a self-supervised manner and then partially fine-tuned on the initial batch of target data. 
Our finding echoes those reported in recent comparative studies of the two types of neural architectures which conclude that there is no absolute winner~\cite{pinto2022impartial,wang2022can}.
The implication for EFCIL is that the use of both types of architecture should be explored in future works.


\section{Conclusion}
We perform an analysis of EFCIL in an evaluation setting that includes numerous and diverse classification tasks.
We confirm the findings of existing comparative studies which have shown that no CIL algorithm is the best in all cases~\cite{belouadah2021_study,masana2021_study,Feillet_2023_WACV} and that algorithms based on transfer learning provide accuracy and stability for EFCIL~\cite{hayes2020_deepslda,janson2022simple}.
Our main finding is that the initial training strategy is the dominant factor influencing the average incremental accuracy, but that the choice of CIL algorithm is more important in preventing forgetting.
Beyond the fact that there is no silver bullet approach to dealing with EFCIL, our in-depth statistical study quantifies the effect of different components of EFCIL approaches and thus enables informed decisions when designing new methods or implementing EFCIL in practice.

\noindent \textbf{Acknowledgements}.
This work was supported by the European Commission under European Horizon 2020 Programme, grant number 951911 - AI4Media. It was made possible by the use of the FactoryIA supercomputer, financially supported by the Ile-de-France Regional Council.

{\small
\bibliographystyle{ieee_fullname}
\bibliography{egbib}
}

\end{document}